\documentclass{article}

\usepackage[english]{babel}

\usepackage[letterpaper,top=2cm,bottom=2cm,left=3cm,right=3cm,marginparwidth=1.75cm]{geometry}

\usepackage{amsmath}
\usepackage{graphicx}
\usepackage[colorlinks=true, allcolors=blue]{hyperref}

\title{Prompt-Engineering and Transformer-based Question Generation and Evaluation}
\author{Rubaba Amyeen}

\begin{document}
\maketitle

\begin{abstract}
Question generation has numerous applications in the educational context. Question generation can prove helpful for students when reviewing content and testing themselves. Furthermore, a question-generation model can aid teachers by lessening the burden of creating assessments and other practice material. This paper aims to find the best method to generate questions from textual data through a transformer model and prompt engineering. In this research, we finetuned a pre-trained distilBERT model on the SQuAD question-answering dataset to generate questions. In addition to training a transformer model, prompt engineering was applied to generate questions effectively using Meta’s LLaMa model. The generated questions were compared against the baseline questions in the SQuAD dataset to evaluate the effectiveness of four different prompts. All four prompts demonstrated over 60\% similarity on average. Of the prompt-generated questions, 30\% achieved a high similarity score greater than 70\%. 

\end{abstract}

\section{Introduction}

Question generation is an essential NLP task. One of its primary applications is that it can be used as a learning tool. Active recall is the practice of retrieving information by answering questions. According to the Association for Psychological Science (APS), active recall is a high-utility learning technique since learners demonstrate higher academic performance after studying with active recall \cite{dunlosky2013improving}. A model that can generate questions would help students use active recall to review material and lessen the time it takes to create questions. According to a study from CBE Life Science Education, students who used active learning strategies scored 5.5\% and 10.2\%  higher than students who did not use active learning strategies on exams 1 and 2, respectively \cite{walck2021extent}. Question generation systems can also motivate students to engage in educational activities because learners will know they will be tested on the material \cite{heilman2011automatic}. Online courses and materials need many questions, so this model would reduce question generation costs \cite{pan2019recent}. Furthermore, it would allow students to test their knowledge on materials that do not include guiding questions. On the other hand, question generation is also valuable for teachers by allowing educators to spend their time elsewhere from formulating questions. Assessments need a constant flow of new questions as the value of questions decreases after multiple usages since test takers may share them.

Question generation is a generative task that involves supervised learning with language data. The task concerns taking the input data (natural language texts of the context and answer) and transforming it into the output data, which is the question. We tested two models: The transformer model distilBERT and Meta’s LLaMA large language model (LLM) through prompt engineering. We evaluated the transformer model using the F1 score and the LLaMA using spaCy vector similarity.

\section{Background}

The task of question generation has rapidly developed over the past few years. Ali and Chali train their model using the TREC 2007 Question Answering Track. The dataset has a series of factoids, lists, and other questions under targets \cite{ali2010automatic}. They simplified complex sentences to elementary structures with syntactic parsing. However, they were unable to address word disambiguation with semantic information. Yuan and Wang implemented a recurrent model to generate questions using the SQuAD dataset \cite{yuan2017machine}. In their implementation, the encoder processes the answer and document while the decoder generates the question. They faced challenges with similar entities, related verbs being swapped, and their model needing common sense.
Similarly, Duan and Tang used a convolution neural network (CNN) and recurrent neural network (RNN) \cite{duan2017question}. Chen and Wu recently implemented a reinforcement learning-based Graph2Sequence model \cite{chen2019reinforcement}. 

Most prior works often generate poor-quality questions. We propose using large language models (LLMs) with prompt engineering to combat these challenges. The prior papers have not addressed a prompt engineering-based approach for the question generation task. Furthermore, as LLMs have been trained on large amounts of data, they have common sense knowledge that prior models lack.







\section{Dataset}

In this paper, Stanford’s SQuAD dataset was used to generate questions. \cite{rajpurkar2016squad}, \cite{rajpurkar2018know}. The transformer model was evaluated and trained on the SQuAD dataset. Since the SQuAD version 2 dataset has unanswerable questions, we used the SQuAD 1.1 dataset \cite{stanford2022squad}. For the question generation task, unanswerable questions would not be useful as we wanted to generate questions that test reading comprehension. This dataset contains 100,000 questions from over 500 articles. While this dataset is commonly used for answering questions, we reversed it for question generation. The SQuAD dataset was split approximately 90/10 between the training set (87599 samples) and the validation set (10570 samples). For data preprocessing, the inputs were tokenized. Long contexts are split; however, in case the answer is where the context is split, the hyperparameter, doc\_strides, allows overlap between the two parts of the split context. We also used the SQuAD dataset for prompt engineering. We compared the SQuAD questions to the prompt-generated questions by LLaMA to evaluate the quality of the generated questions.

We visualized the SQuAD dataset with simple techniques, such as a histogram of the length of questions with outliers accounted for and the most frequent keywords with stop words removed.

\begin{figure}[hbt!]
\centering
\includegraphics[width=0.45\textwidth]{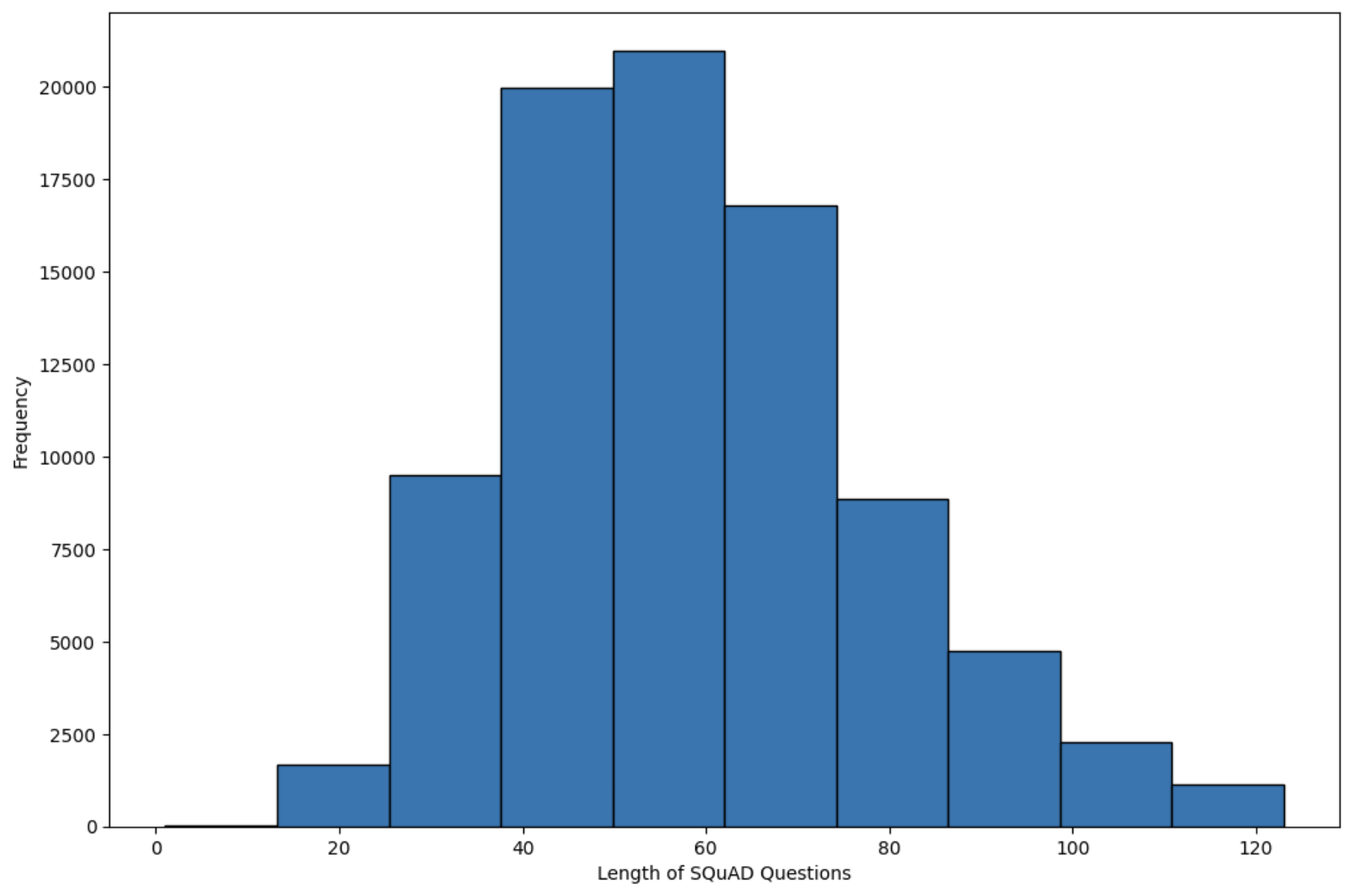}
\caption{\label{fig:squad}Length of the Questions in SQuAD.}
\end{figure}

\begin{figure}[hbt!]
\centering
\includegraphics[width=0.45\textwidth]{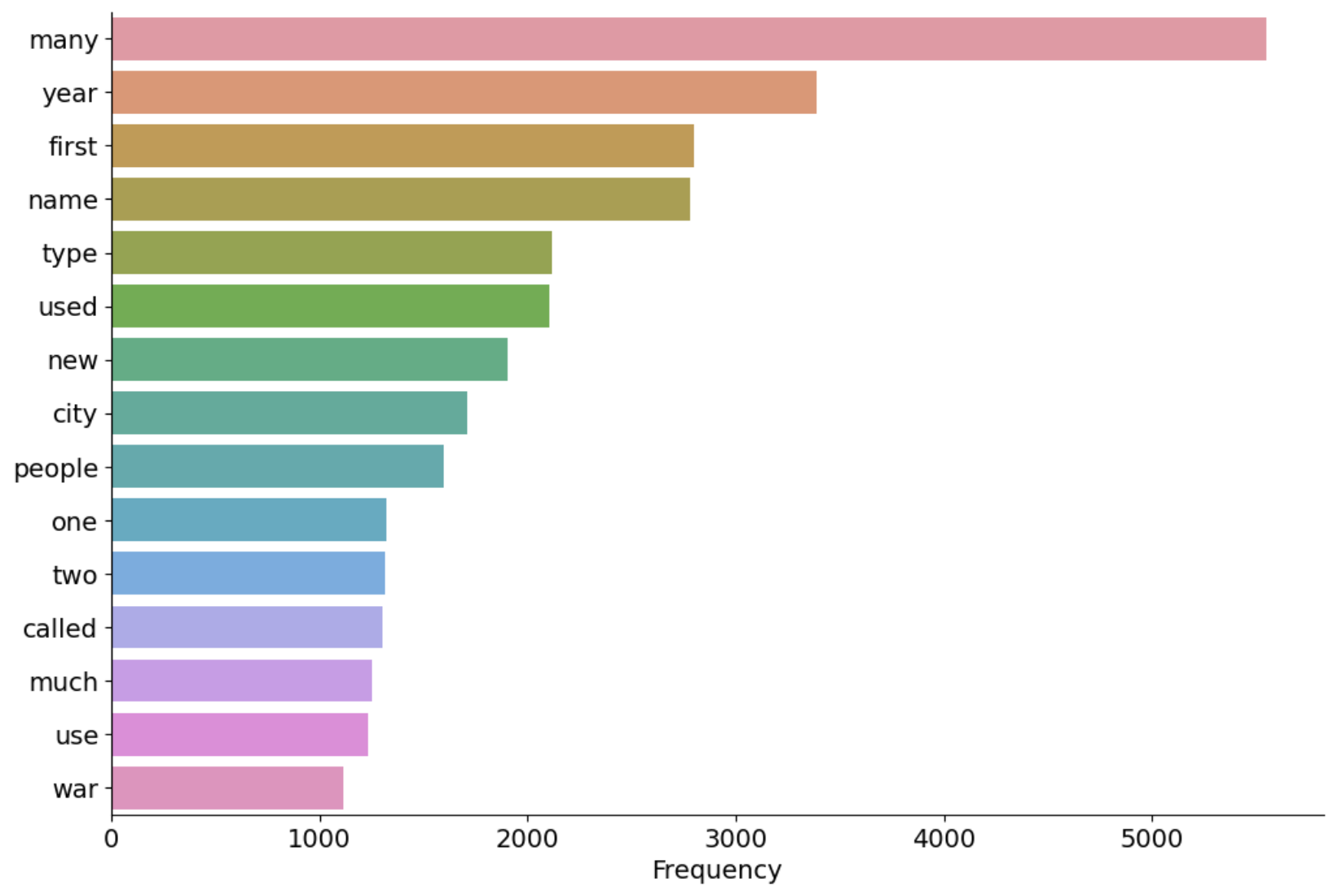}
\caption{\label{fig:squadfrequent}Most Frequent Words in SQuAD Questions.}
\end{figure}




\section{Methodology/Models}


\subsection{Transformer DistilBERT}

Transformers are deep learning models that consist of a unique architecture that retains the relationships between all the words in a sentence. Transformers are not implemented with recurrent neural networks (RNNs) and only use attention mechanisms \cite{vaswani2017attention}. They consist of an encoder and a decoder. An encoder extracts features from the input, and a decoder produces a prediction for the task using the features. BERT (Bi-directional Encoder Representation from Transformers) is a widely used pre-trained language model in deep neural networks \cite{devlin2018bert}. Figure 3 illustrates the distilBERT model architecture, a smaller version of the BERT model.  

\begin{figure}[hbt!]
\centering
\includegraphics[width=0.4\textwidth]{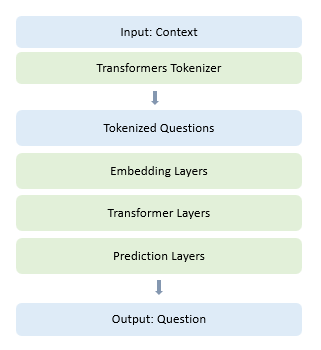}
\caption{\label{fig:distil}DistillBERT model architecture.}
\end{figure}

DistilBERT is a distilled version of BERT with 40\% fewer hyperparameters than BERT-base-uncased \cite{sanh2019distilbert}. We finetuned the DistilBERT model on the SQuAD dataset for question generation. First, the SQuAD dataset was reversed, that is, we transformed the dataset into Pandas Dataframe and swapped the “question” and “answer” columns. 

The parameters to finetune the pre-trained model included a learning rate of .00005, weight decay of 0.01, and training epochs set to 3. Figure 4 displays the procedure for the distilBERT question generation task. 

\begin{figure}[hbt!]
\begin{center}
\includegraphics[width=0.5\textwidth]{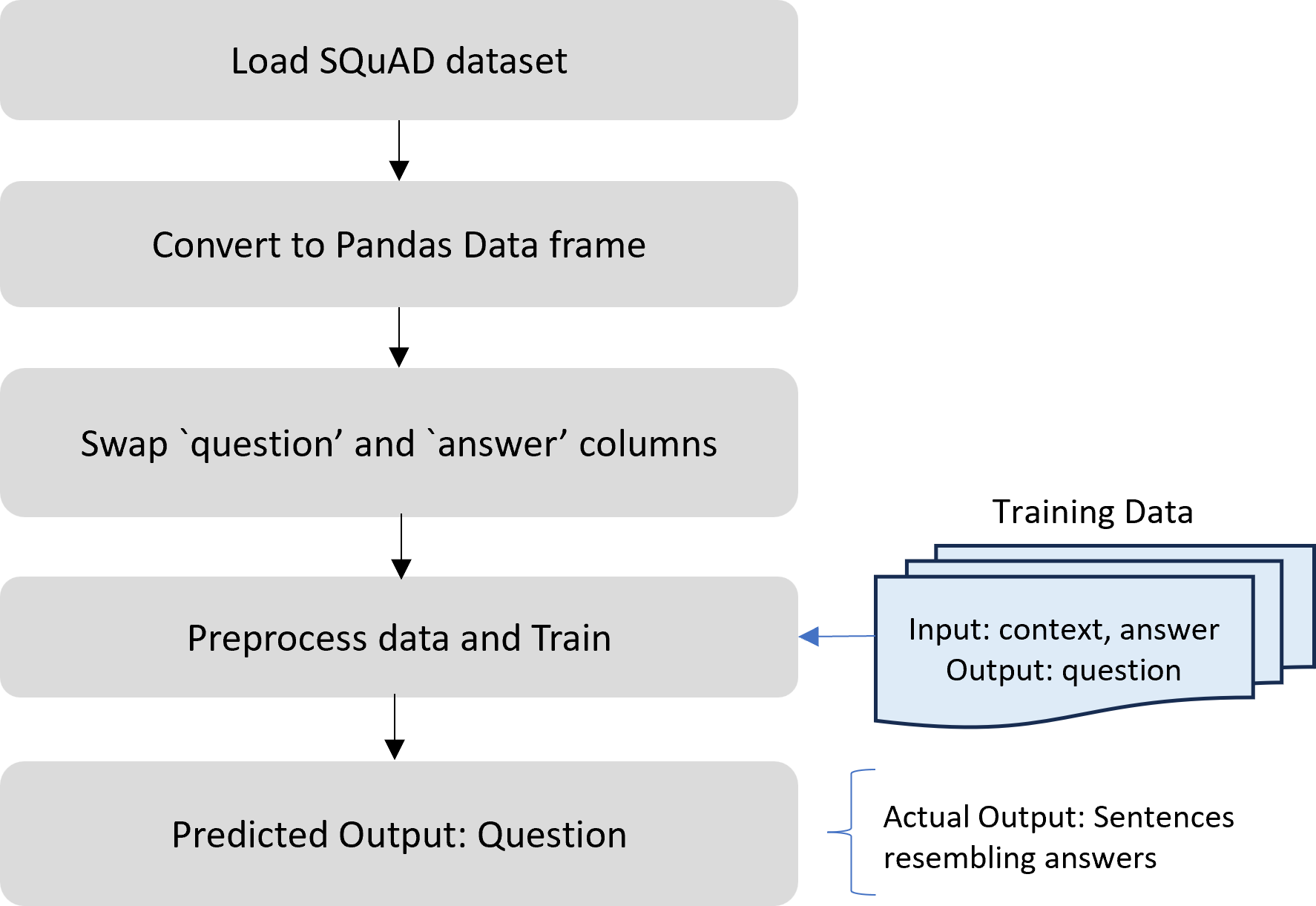}
\caption{\label{fig:transformer} Transformer question generation flow}
\end{center}
\end{figure}

\subsection{Prompt Engineering}

Prompt engineering is the practice of formulating specific prompts for LLMs to generate desired outputs. Andrew Ng and Isa Fulford outlined two main principles for prompt engineering: writing clear and specific instructions and allowing the model to think \cite{prompteng}. We developed four prompts for the question generation task using these guidelines for the Meta's LLaMA model \cite{touvron2023llama}. The temperature, the prompts' variability, was set to 0.5 to ensure varied responses. Each of the prompts generated five questions. Table 1 shows the four prompts. 

\begin{table}[hbt!]
\vfill
\begin{center}   
\begin{tabular}{p{0.15\linewidth} | p{0.7\linewidth}}
Prompt & Description \\\hline
Prompt-A & ``Generate 5 questions from the text;"  \\
Prompt-B & ``Generate 5 complex questions from the text." \\
Prompt-C & ``Generate 5 questions from the text; make sure the questions can be answered." \\
Prompt-D & ``Generate 5 questions from the text; answer the question in the text; if the question is answered in the context, output 5 questions." \\
\hline
\end{tabular}
\caption{\label{tab:prompts} Written Prompts.}
\end{center}
\vfill
\end{table}

We randomly selected a sample of 50 contexts from the SQuAD dataset. We evaluated the four distinct prompts against the baseline questions under each context from SQuAD. For each prompt, 250 questions were generated. We used the spaCy metric that used word embeddings to measure the sentence similarities between the generated questions and baseline questions. We recorded the max similarity score for each generated question and the max similarity score for each prompt type under every context. Figure 5 shows the process of prompt engineering and how the prompts were evaluated. 

\begin{figure}[hbt!]
\begin{center}
\includegraphics[width=0.6\textwidth,height=3in]{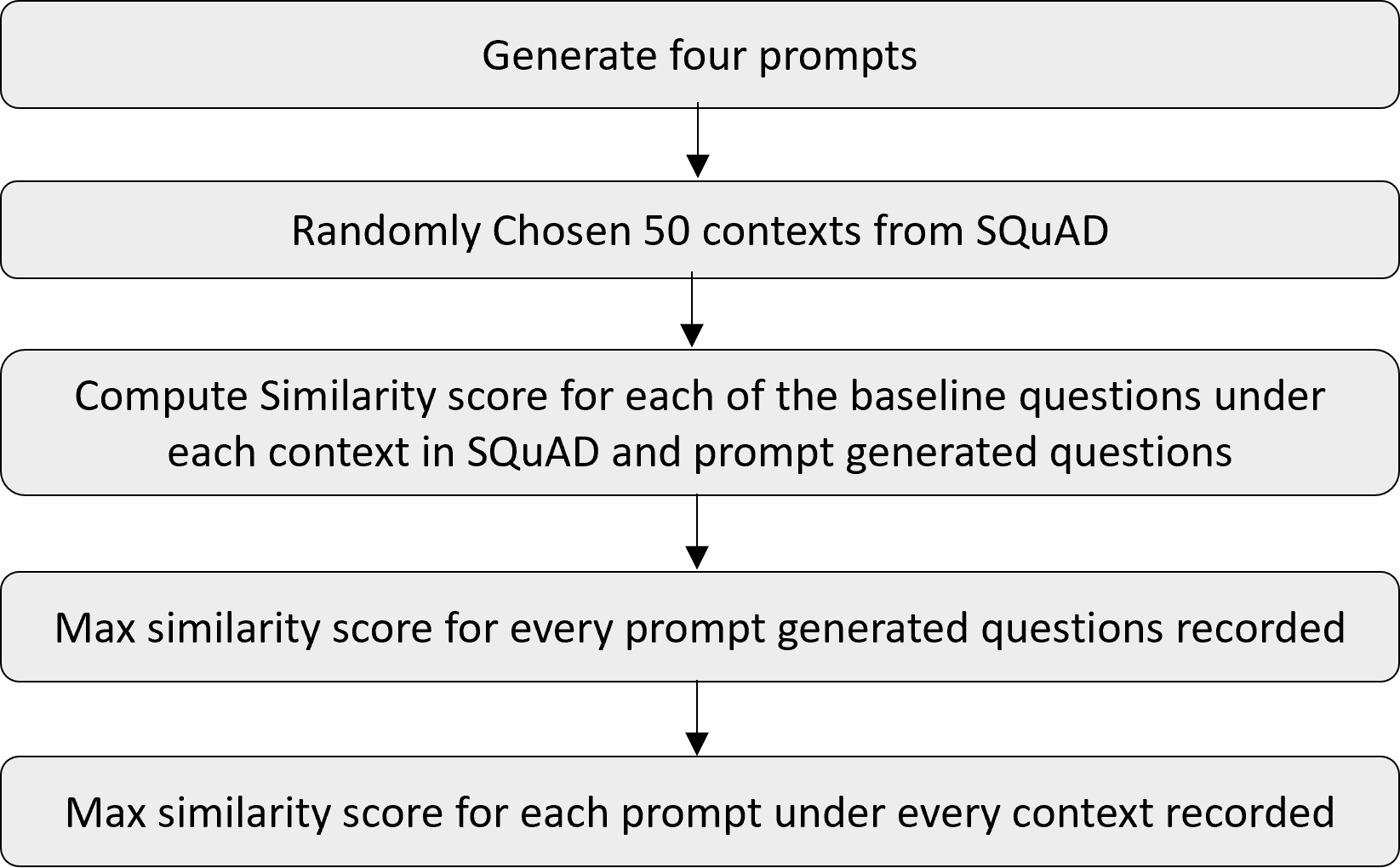}
\caption{\label{fig:flow} Flow Diagram of Data Collection Process}
\end{center}
\end{figure}

Table 2 is a sample of the data collection for the context: ``Beyonc\'e Giselle Knowles-Carter (born September 4, 1981) is an American singer, songwriter, record producer and actress. Born and raised in Houston, Texas, she performed in various singing and dancing competitions as a child, and rose to fame in the late 1990s as lead singer of R\&B girl-group Destiny\'s Child. Managed by her father, Mathew Knowles, the group became one of the world\'s best-selling girl groups of all time. Their hiatus saw the release of Beyonc\'e\'s debut album, Dangerously in Love (2003), which established her as a solo artist worldwide, earned five Grammy Awards and featured the Billboard Hot 100 number-one singles `Crazy in Love' and `Baby Boy'." The generated questions from the four prompts are displayed. The ``Prompt Max" is the max similarity score for all the generated questions under the specific prompt.

\begin{table}[t!]
\begin{center}   
\begin{tabular}{l|c| p{0.5\linewidth}|p{0.1\linewidth}|p{0.09\linewidth}}
\hline
Context & Prompt & Generated Question & Question Max & Prompt Max \\\hline
Beyonc\'e ... & A & What is Beyonc\'e's full name? & 0.833 & 0.935  \\
 & A & When was Beyonc\'e born? & 0.534 &  \\
 & A & Where did Beyonc\'e grow up? & 0.749 &  \\
 & A & Who managed Destiny's Child? & 0.857 \\
 & A & What was the name of Beyonc\'e's debut album? & 0.935 \\
 & B & What is Beyonc\'e's full name? & 0.833 & 0.833 \\
 & B & Where was she born and raised? & 0.472 \\
 & B & In what decade did she rise to fame as the lead singer of Destiny's Child? & 0.722 \\
 & B & Who managed her career, her father Mathew Knowles or someone else? & 0.433 \\
 & B & Which album marked her debut as a solo artist worldwide and earned five Grammy Awards? & 0.639 \\
 & C & What is Beyonc\'e's first name? & 0.865 & 0.865  \\
 & C & When was Beyonc\'e born? & 0.534 &  \\
 & C & Where did Beyonc\'e grow up? & 0.749 &  \\
 & C & Who managed Destiny's Child? & 0.857 \\
 & C & Which album established Beyonc\'e as a solo artist worldwide? & 0.719 \\
 & D & What is Beyonc\'e's birth name? & 0.879 & 0.956 \\
 & D & When was Beyonc\'e born? & 0.535 \\
 & D & Where did Beyonc\'e grow up? & 0.749 \\
 & D & How did Beyonc\'e rise to fame? & 0.956 \\
 & D & What was the name of Beyonc\'e's debut solo album? & 0.934 \\
 
\hline
\end{tabular}
\caption{\label{tab:sample} Data Collection Table for the context $1$}
\end{center}
\end{table}

\section{Results and Discussion}

\subsection{Transformer Model}

Using the evaluation metric under the Trainer class from PyTorch, the F1 score for the transformer model was 15.88. A key reason for a low F1 score was found to be that the transformer model could not generate questions and instead generated sentences similar to the answers in SQuAD. These predicted outputs may be because the model was already familiar with question-answering tasks and generating answers, so it could not generate questions. The distilBERT model was trained on 3.3 billion words from the Wikipedia and Toronto Book Corpus, so finetuning on a much smaller dataset like SQuAD was not effective in reversing distilBERT’s inherent behavior to answer questions. As the transformer model was not able to generate questions, we did not use a metric to evaluate the generated outputs.

\subsection{Prompt Engineering}

\begin{figure}[hbt!]
\begin{center}
\includegraphics[width=0.6\textwidth]{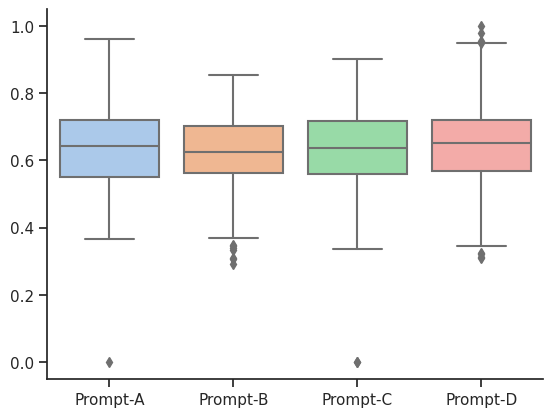}
\caption{\label{fig:distribution} Score distribution across diffent prompts}
\end{center}
\end{figure}

We used the spaCy metric for similarity scoring. Figure 6 is a boxplot of the similarity scores for all the generated questions under each prompt. The boxplot reveals that the median similarity score for Prompt D’s generated questions was the highest at 0.6444. Furthermore, Prompt D also had the highest outliers, with one being a perfect match from the similarity score of 1. Prompts A and C had an average similarity score of 0.6387 and 0.6321 respectively. Prompt B performed the worst, with the lowest average of 0.6227. 

\begin{figure}[hbt!]
\begin{center}
\includegraphics[width=0.6\textwidth]{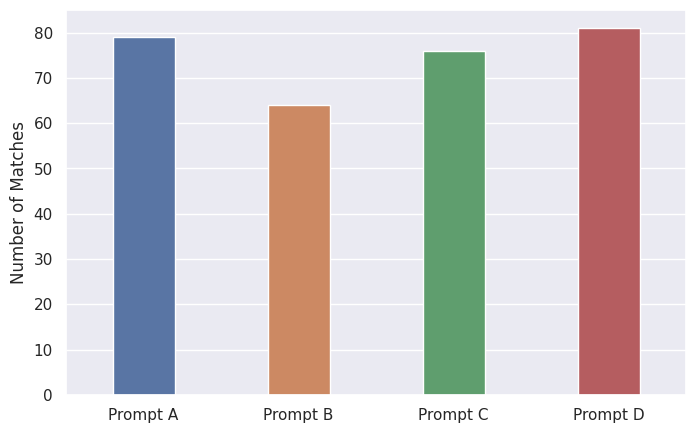}
\caption{\label{fig:matches} High Threshold Score Distribution for the Prompts}
\end{center}
\end{figure}

Another method we used to establish which prompt performed the best was the prompt that had the most matches with the baseline questions. We established a threshold of 0.7 similarity score to be considered a match. After comparing generated questions with a similarity score above 0.7 with the baseline questions, we discovered that most of the generated questions had the same meaning as the baseline question; while generated questions with a similarity score below 0.7 were inherently different. We defined a match as when the baseline and generated questions were nearly identical. Figure 7 reveals that Prompts A, B, C, and D had 79, 64, 76, and 81 matches, respectively, out of the 250 questions each prompt generated in total. Prompt D had the highest number of matches, so it could generate the highest quality questions, assuming the SQuAD dataset has the highest quality questions. Prompt B had the lowest number of matches. The word “complex” in Prompt B may have generated unanswerable questions from the context and required more critical thinking, so it performed poorly under the similarity metric. Table 3 displays an unanswerable question generated by Prompt B.

\begin{table}[hbt!]
\vfill
\begin{center}   
\begin{tabular}{p{0.7\linewidth} | p{0.2\linewidth}}
Context & Prompt-B Question \\\hline
The Hanover Zoo is one of the most spectacular and best zoos in Europe. The zoo received the Park Scout Award for the fourth year running in 2009/10, placing it among the best zoos in Germany. 
 & Where is the Hanover Zoo located in Germany?  \\

\hline
\end{tabular}
\caption{\label{tab:promptb} Sample Unanswerable Question from Prompt B.}
\end{center}
\vfill
\end{table}

\pagebreak
Lastly, Figure 8 is a distribution graph of the max scores of every prompt under each of the 50 contexts in total. Generally, the red (Prompt D) and blue (Prompt A) lines are more significant than the orange (Prompt B) and green (Prompt C) lines. Prompt B had a particularly low max similarity score for context 45 of 0.42 compared to Prompt A, C, and D, which had max similarity scores of 0.73, 0.70, and 0.77, respectively. This is most likely because Prompt B generated compound questions, where two questions were combined into one. For example, one of the questions was, “In what way can Nirvana be seen as a liberation from the cycle of suffering, and how is this achieved?”

\begin{figure}[hbt!]
\begin{center}
\includegraphics[width=0.7\textwidth]{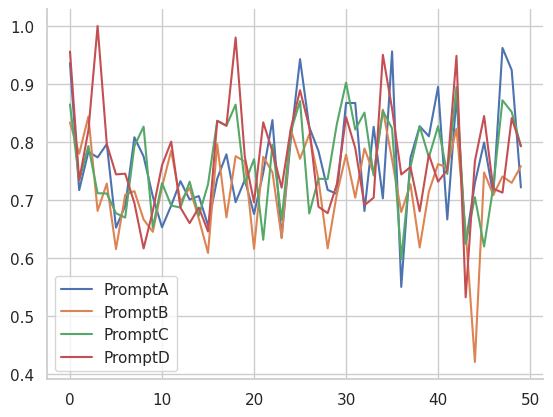}
\caption{\label{fig:max} Max score distribution across contexts}
\end{center}
\end{figure}

\section{Conclusion}

In this paper, we finetuned a distilBERT model on the question generation task. This transformer model could not generate questions with an F1 score of 15.88. On the other hand, Meta’s LLaMA model successfully generated questions using the prompts we created with the prompt engineering guidelines. As the transformer model was not able to generate questions and Meta’s LLaMA model was, we decided to only evaluate the generated questions from LLaMA. We used a similarity metric to determine which of the four prompts we created was the most effective. We found that the most complex prompt (Prompt D) resulted in the most matches with the baseline questions from SQuAD at 81 matches. The most simple prompt (Prompt A) performed the second best with 79 matches. Prompt C had 76 matches. Lastly, Prompt B had a significantly lower number of matches at 64 matches. It should be noted that Prompt B was the same as Prompt A except for the added word “complex,” suggesting that the similarity score metric rewards simple factual questions rather than compound questions.  

We plan to extend our research to include other LLMs, such as the Dolly model, GPT-4, and PaLM 2. Furthermore, we plan to combine the question generation model with a classification model that classifies the contexts under specific educational subjects to make the questions generated more appropriate for each subject beyond reading comprehension.

\section{Acknowledgements}
We would like to acknowledge the technical contributions and mentorship of Tanish Jain from Stanford University.


\bibliographystyle{alpha}
\bibliography{sample}
\end{document}